%% file: PromptingTowards_Main.tex
\documentclass[10pt, a4paper]{article}

\usepackage{lrec-coling2024} 

\usepackage{tikz}
\usepackage{amsmath}
\usepackage{caption}
\usepackage{subcaption}
\usepackage{tabularx}
\usepackage{booktabs} 
\usepackage{multirow} 
\usepackage{amssymb} 
\usepackage{pifont} 
\newcommand{\xmark}{\ding{55}}
\newcommand{\ii}[1]{{\footnotesize \textcolor{gray}{#1}}}

\title{Prompting Towards Alleviating Code-Switched Data Scarcity in Under-Resourced Languages with GPT as a Pivot}
		
\name{Michelle Terblanche, Kayode Olaleye, Vukosi Marivate} 

\address{Data Science for Social Impact\\Dept. of Computer Science\\University of Pretoria\\
	South Africa \\
	michelle.terblanche@gmail.com,\\ kayode.olaleye@cs.up.ac.za, \\vukosi.marivate@cs.up.ac.za \\}

\abstract{
	Many multilingual communities, including numerous in Africa, frequently engage in code-switching during conversations. This behaviour stresses the need for natural language processing technologies adept at processing code-switched text. However, data scarcity, particularly in African languages, poses a significant challenge, as many are low-resourced and under-represented. In this study, we prompted GPT 3.5 to generate Afrikaans–English and Yoruba–English code-switched sentences, enhancing diversity using topic-keyword pairs, linguistic guidelines, and few-shot examples. Our findings indicate that the quality of generated sentences for languages using non-Latin scripts, like Yoruba, is considerably lower when compared with the high \mbox{Afrikaans--English} success rate. There is therefore a notable opportunity to refine prompting guidelines to yield sentences suitable for the fine-tuning of language models. We propose a framework for augmenting the diversity of synthetically generated code-switched data using GPT and propose leveraging this technology to mitigate data scarcity in low-resourced languages, underscoring the essential role of native speakers in this process.
    \\ \newline \Keywords{code-switch, LLM, few-shot, prompting} }

\begin{document}

\maketitleabstract

\input{introduction}
\input{related_work}
\input{exp_setup}

\input{evaluation}
\input{conclusions}

\nocite{*}
\section{Bibliographical References}\label{sec:reference}
\bibliographystyle{lrec-coling2024-natbib}
\bibliography{refs}

\end{document}

%% file: introduction.tex
\section{Introduction}
\label{sec:introduction}

Multilingual communities, exemplified well by various African countries, often engage in code-switching, where two or more languages are used within a single discourse \cite{PoplackCS}. This language practice highlights the need to develop more advanced natural language processing (NLP) technologies that can smoothly process and produce code-switched sentences. This will move the needle towards equitable representation of the world’s under-resourced languages, ensuring that everyone has equal access to these technologies \citep{solorio_needle}.

There are numerous challenges in code-switching research. The main three are highlighted by \citet{Do2021} as follows: i) data, which is related to quantity, quality and availability; ii) evaluation, which refers to benchmarks and metrics; and iii) challenges related to end-to-end applications, particularly the ability to process and produce code-switched data.

The focus of this paper is on the first challenge regarding \textit{\textbf{data}}. While code-switching frequently occurs in written forms, due to the ubiquitous use of social media platforms, leveraging this data in NLP applications for code-switching presents many challenges. These platforms, with their extensive and diverse linguistic expressions, can be invaluable in gathering code-switched data. Yet, the practical utility of such data is hindered by various factors, including the informal, inconsistent nature of online language \citep{Cetinoglu2016}. It is common to use acronyms, emojis and make spelling mistakes which affect quality and usability of such data \mbox{\citep{Srivastava2019}}.  Furthermore the diversity of such data is limited to a specific type of language use \citep{Winata2022}.

To address the shortage of available data, efforts have been made to create synthetic code-switched data using different methods: from using parallel corpora with linguistic constraints on where a switch can occur \citep{Pratapa2018,Rizvi2021} to employing transformer-based models to generate diverse sentences that adhere to lexical and syntactic rules \citep{Raghuveer2022}. A more recent study evaluated prompting of large language models (LLMs) to generate code-switched data for South East Asian languages \citep{Yong2023}. They explored a few prompting templates with a limited number of topics in a zero-shot manner and cautioned against the use of synthetically generated data without involving native speakers of the language.

In this paper, we build on the work of \citep{Yong2023} to address the question about \textit{GPT's} ability to generate code-switched data. Our work overlaps in that we also use an LLM, \textit{OpenAI's GPT}, and various topics in the prompts. We increase the number of topics and provide topic-related keywords in an effort to increase diversity and reduce the model's propensity to default to certain words. Our goal is not to evaluate various prompting templates, however, we add linguistic guidelines in the prompts to further increase diversity. We propose this as an approach towards language agnostic prompting. We also test the performance of GPT 3.5 with few-shot in-context examples. We specifically consider whether \textit{GPT} can support the generation of larger code-switched datasets and to what extent.

Our contributions are as follows: (i) we provide a framework to increase the diversity of synthetically generated code-switched data by prompting \textit{OpenAI's GPT}; and (ii) we position GPT as a pivot to address code-switched data scarcity in low-resource languages while emphasising the need for native speakers in the loop. 
	
Increasing data availability is at the center of developing language models that serve multilingual communities. Our work is a step towards closing the gap in low-resourced and under-represented languages.

%% file: related_work.tex
\section{Related Work}
\label{sec:rel_work}
\subsection{Code-Switching Research}
\label{sec:rel_work:csresearch}

Various types of code-switching have been identified but the type that attracts the most academic research is intra-sentential code-switching which can occur anywhere within a sentence boundary \citep{Poplack1980} and as a result, adds complexity in evaluation \citep{Poplack2001}. Another complex type is intra-word code-switching where the stem of one language is bound to another language \citep{Cetinoglu2016,van-der-westhuizen-niesler-2018-first}.  

Over and above the issue of data diversity \citep{Winata2022}, one of the major challenges in code-switching studies is related to data availability \citep{Do2021}. A survey by \citep{Winata2022} showed that up until October 2022, a relatively small amount of papers (\citealp{ACLAnthology} and \citealp{ISCAProceedings}) focused on code-switching research in African languages with very few publicly available datasets. Eleven publications mention South African languages. The non-English South African languages referenced are isiZulu, isiXhosa, Setswana, Sesotho and Afrikaans. Only one proceeding includes Afrikaans code-switching \citep{niesler2008accent} with no published dataset. A paper by \citet{van-der-westhuizen-niesler-2018-first} introduced the first corpus on isiZulu, isiXhosa, Setswana, Sesotho curated from transcribed soap opera speech data and eight of the papers makes use of this dataset and is mainly focused on automatic speech recognition (ASR) systems.

Code-switching in Kiswahili--English is studied in two papers but no datasets were made available \citep{otundo2022intonation,piergallini2016word}. In addition to a survey by \citet{Winata2022}, one other paper was found that addresses Sepedi--English code-switching. \citet{modipa2013implications} develop a corpus from a set of radio broadcasts to evaluate the implication of code-switching in ASR systems. This dataset is publicly available.    
This brief review of the state of code-switching research in an African context motivates our work to develop methods for addressing data scarcity.

A predominant approach to mitigating data availability issues involves augmenting existing datasets through the generation of synthetic code-switched data. Some of the methods to augment the earlier mentioned South African speech corpus include the use of word embeddings to synthesise code-switched bigrams to find similar words in the sparse training data \citep{Westhuizen2017}. \citet{Biswas2018} evaluated adding out-of-domain monolingual data and synthesised code-switched data using an LSTM to augment the dataset.

For non-African languages, \citet{Rizvi2021} developed a toolkit that generates multiple code-switched sentences using either the Equivalence Constraint or the Matrix Language Frame. The limitations are that it relies on a good sentence aligner and parser and parallel translated sentences as input. The notion is that this approach should work on any language pair. \citet{winata2019codeswitched} implemented a sequence-to-sequence model for English-Mandarin code-switched data. Although the model does not require external knowledge regarding word alignments, it still relies on an existing English--Mandarin code-switched dataset and parallel corpora. The work of \citep{liu2020attention} introduced an attention-informed zero-shot adaptation method that relies on a limited number of parallel word pairs. The languages covered are German, Italian, Spanish and Thai, the latter two for natural language understanding. The shortcoming of the above-mentioned approaches is the diversity of data. Most existing code-switched datasets were collected from social media platforms such as Twitter and therefore limits the type of code-switching \citep{Do2021}.

To this issue, \citet{Raghuveer2022} developed an encoder-decoder translation model for controlled code-switched generation. It uses monolingual Hindi and a publicly available Hindi--English code-switched dataset as input to generate data that is faithful to syntactic and lexical attributes.

\citet{Yong2023} proposed an approach that is independent of existing code-switched datasets or parallel corpora through prompting of LLMs.
Their objective was to test whether multilingual LLMs can generate code-switched text through prompting. They evaluated a variety of prompt templates and found that those explicitly defining code-switching gave the highest success rate. However, they also highlighted the sentences often contained word-choice errors and semantic inaccuracies which was more prevalent in the languages that don't use the English alphabet and Latin script. They limited the scope to five topics and did not include diversity as a measure. Their findings were that GPT's capability to generate code-switched data is superior to other LLMs, however, using this method without humans-in-the-loop is not advised.

\citet{susmit} elaborated on LLMs such as GPT being prone to hallucinations where it provides factually inaccurate or contextually inappropriate responses. A solution to address this is to ensure carefully curated prompts. Furthermore, to avoid encoded biases, \citet{Bender2021} emphasises the need to also evaluate appropriateness in relation to a particular social context.

With the rapid adoption of LLMs in everyday life, these are a low-cost alternative to alleviate data scarcity in low-resourced and under-represented languages by synthetically generating text. In this paper we expand on the work of \citet{Yong2023} and position GPT as a pivot in generating code-switched data rather than a self-sufficient solution.

%% file: exp_setup.tex
\section{Code-Switched Text Generation via GPT-3.5 Prompting}
\label{sec:method}

Our prompt-based approach to code-switched (CS) text generation is heavily inspired by the work of \citet{Yong2023}, who collected synthetic CS data by prompting LLMs with requests along languages and topics. Their focus was on code-switching English with South-East Asian languages. In our case, we focus on two under-explored and under-resourced code-switching scenarios: Afrikaans--English and Yoruba--English. Although Afrikaans and English are typologically dissimilar \citep{vandulm}, they are both West Germanic languages and generating CS text should be easier. Yoruba is a tonal language and even more dissimilar to English which could provide challenges when creating synthetic CS data. We extend the limited topics covered in \citet{Yong2023} and present GPT-3.5 not as an autonomous solution to CS data scarcity, but as a potential tool for supporting CS data curation efforts for under-resourced African languages. We specifically use GPT-3.5, firstly as a baseline to compare with the findings from \citet{Yong2023} and secondly, due to the unavailability of the GPT-4 API at the time of our experiments\footnote{The API for GPT 4 was made available after we finished the majority of the experiments, \newline\url{https://openai.com/blog/gpt-4-api-general-availability}}.

\subsection{Prompting for Afrikaans--English CS Sentences}
\label{sec:dataset:prompt}
Building on the prompt template from \citet{Yong2023}, which uses topics as guidelines, our approach extends this by (i) incorporating specific code-switching words related to each topic within the prompt and (ii) evaluating the effect of prompt complexity from basic (Section~\ref{sec:dataset:prompt:gen}) to more comprehensive prompts (Section~\ref{sec:dataset:prompt:ling}).

We curate a non-exhaustive list of common conversation topics and associate typical English words from native speakers of Afrikaans and from available online platforms. We cover 22 topics with a total of 355 keywords. For this paper we generate one sentence per keyword for the various prompts. We also develop a general list of words used in code-switching that is not directly linked to a specific topic consisting of 138 words. $\sim$90\% of the keywords are nouns, verbs and adjectives which is in line with the notion that switching is more likely to occur on these open word classes as opposed to close word classes (such as pronouns and conjunctions) \citep{kodali2022symcom}.

\subsubsection{Topic-Keyword Basic Prompting}
\label{sec:dataset:prompt:gen}

In the six different prompting templates of \citet{Yong2023}, one prompt specifically requests a native speaker to give a mixed sentence. This is an indirect way to impose a matrix language (ML). We explicitly include the use of a matrix language in our prompts \citep{jake2002making}. This is to ensure that we adequately represent the low-resourced language. However, we recognise that grammatical constraints on CS is an open research question with varying definitions of acceptability that evolves over time \citep{Bhat2016}.

The following shows the basic prompt we used (Prompt 1.1) and a few examples to highlight the behaviour of GPT-3.5 (English translation in \textit{Italics}).

\hspace{-4mm}
	\begin{tikzpicture}
		
		\node[text width=0.97\columnwidth, rounded corners, fill=gray!10, draw=none, anchor=north] {
			\textbf{Prompt 1.1:} Generate an Afrikaans-English code-switch sentence with Afrikaans as the matrix language. Typical words used in code-switching are: \textbf{general}\footnotemark. The topic is \textit{[insert topic]} and must contain the word \textit{[insert keyword]}.
		};
		
	\end{tikzpicture}
	\footnotetext{List of general words provided}
	
\textbf{Topic}: education and training; \textbf{keyword}: skills\\[1ex]
\begin{tikzpicture}
	\node[text width=0.97\columnwidth, rounded corners, fill=blue!10, draw=none, anchor=north] {
	\textbf{Example 1}: Ek$_{\text{af}}$ moet$_{\text{af}}$ my$_{\text{af}}$ skills$_{\text{en}}$ verbeter$_{\text{af}}$ om$_{\text{af}}$ 'n$_{\text{af}}$ beter$_{\text{af}}$ werksgeleentheid$_{\text{af}}$ te$_{\text{af}}$ kry$_{\text{af}}$.
	
	\textit{I must improve my skills to get a better job opportunity.}
	};
\end{tikzpicture}

\vspace{1mm}
\textbf{Topic}: general conversation; \textbf{keyword}: try\\[1ex]
\begin{tikzpicture}
	\node[text width=0.97\columnwidth, rounded corners, fill=blue!10, draw=none, anchor=north] {
		\textbf{Example 2}: Ek$_{\text{af}}$ sal$_{\text{af}}$ probeer$_{\text{af}}$ to$_{\text{en}}$ finish$_{\text{en}}$ my$_{\text{af}}$ assignment$_{\text{en}}$ op$_{\text{af}}$ tyd$_{\text{af}}$.
		
		\textit{I will try to finish my assignment on time.}
	};
\end{tikzpicture}

\vspace{0.5mm}

The matrix language is Afrikaans in Example 1 and English in Example 2. We see from these examples that GPT 3.5 does not necessarily follow the prompt with regards to the matrix language. 

We do not evaluate word-level language identification therefore we do not explicitly measure adherence to the matrix language prompt in this paper.

The results of the generated sentences therefore indicate that GPT 3.5 is capable of generating some coherent sentences and can be corrected where the grammatical structure follows English. Section~\ref{sec:evaluation:manual} gives a more detailed analysis of code-switch acceptability.
 
A key observation from using this basic prompt for generating Afrikaans--English sentences is that sentences are one-dimensional with $\sim$80\% of sentences starting with a singular personal pronoun: \textbf{`Ek'} (\textit{English: `I'}) (Section~\ref{sec:evaluation:stats}). This creates the opportunity to explore ways of adding diversity to the type of sentence through the use of basic linguistic guidelines (such as specifying pronouns) which is discussed in the following section.

\subsubsection{Linguistic-Based Prompting}
\label{sec:dataset:prompt:ling}

Since the word lists contain nouns, verbs and adjectives related to specific topics, content diversity in the sentences is addressed. These are also words that are most typically code-switched \citep{kodali2022symcom}. To add further diversity in the type of sentence, we add basic linguistic guidelines in the form of varying pronouns (personal, impersonal, interrogative etc.), tenses (past, present and future that alters the verb) and using negative particles. The inclusion of negative particles is randomly initialised and not in each prompt. We also impose a rule that conjunctions must be in the matrix language since conjunctions are part of closed word classes and should less likely be switched. 

Prompt 2.1 is an example of a prompt using linguistic guidelines following with an example of the generated sentence (English translation in \textit{Italics}). In Example 3 the prompts are adhered to, however, the conjunctions `but' and `and' are in English therefore note adhering to the guideline.
Our preliminary observation is that the prompting approach can support the generation of CS sentences that are diverse. The effect of varying pronouns on sentence diversity is further evaluated in Section~\ref{sec:evaluation:div}. Word order structure mimics that of natural speech and can be corrected where needed. We give additional examples and an evaluation of the quality of the sentences in Section~\ref{sec:evaluation:manual}.

\vspace{2mm}
\hspace{-4mm}
	\begin{tikzpicture}
		
		\node[text width=0.97\columnwidth, rounded corners, fill=gray!10, draw=none, anchor=north] {
			\textbf{Prompt 2.1:} Generate an Afrikaans-English code-switch sentence with Afrikaans as the matrix language. Typical words used in code-switching are: \textbf{general}.
			The topic is \textit{[insert topic]} and must contain the word \textit{[keyword]}.
			Start the sentence with \textit{[insert pronoun]} using the \textit{[insert tense]}. \textit{A conjunction must be Afrikaans.} \textit{[Use a negative particle]}. 
		};
		
	\end{tikzpicture}

	\vspace{1mm}
	\textbf{Topic}: physical health and fitness; \textbf{keyword}: race; \textbf{Pronoun}: impersonal; \textbf{Tense}: past; \textbf{Use a negative particle}: No\\[1ex]
	\begin{tikzpicture}
		\node[text width=0.97\columnwidth, rounded corners, fill=blue!10, 	draw=none, anchor=north] {
			\textbf{Example 3}: Dit$_{\text{af}}$ was$_{\text{af}}$ super$_{\text{en}}$ lekker$_{\text{af}}$ om$_{\text{af}}$ die$_{\text{af}}$ race$_{\text{en}}$ te$_{\text{af}}$ hardloop$_{\text{af}}$, but$_{\text{en}}$ ek$_{\text{af}}$ ignore$_{\text{en}}$ die$_{\text{af}}$ consequences$_{\text{en}}$ and$_{\text{en}}$ het$_{\text{af}}$ te$_{\text{af}}$ veel$_{\text{af}}$ geëet$_{\text{af}}$ afterwards$_{\text{en}}$.
			
			\textit{It was super nice to run the race, but I ignore the consequences and ate too much afterwards.}
		};
	\end{tikzpicture}
	
	\vspace{3mm}

\subsubsection{Few-Shot Prompting}
\label{sec:dataset:prompt:few}

In the work from \cite{Yong2023} they did not evaluate the effect of few-shot examples. We therefore evaluate two additional prompts: Prompt 1.2 and Prompt 2.2 where we add five examples of code-switched sentences to Prompts 1.1 and 2.1 respectively. These are general examples and not in the context of the topic.
	
\subsection{Prompting for Yoruba--English CS Sentences}
\label{sec:dataset:prompty}
	
	In this section we apply the same methodology (Section~\ref{sec:dataset:prompt}) used to generate Afrikaans--English CS sentences to generate Yoruba--English CS sentences and provide brief observations. We develop similar topic keyword lists for Yoruba with most words overlapping with those developed for Afrikaans--English. In future work we will focus on developing common lists that cover a more diverse set of languages. The following are a few examples of the generated Yoruba--English sentences:
	
	\vspace{1mm}
	\textbf{Topic}: information technology; \textbf{keyword}: spreadsheet; \textbf{Pronoun}: indefinite; \textbf{Tense}: future; \textbf{Use negative particle}: Yes\\[1ex]
	\begin{tikzpicture}
		\node[text width=0.97\columnwidth, rounded corners, fill=blue!10, draw=none, anchor=north] {
			\textbf{Example 1}: Mo$_{\text{yo}}$ ni$_{\text{yo}}$ ko$_{\text{yo}}$ relax$_{\text{en}}$, infact$_{\text{en}}$ mo$_{\text{yo}}$ gba$_{\text{yo}}$ surprise$_{\text{en}}$ pe$_{\text{yo}}$ spreadsheet$_{\text{en}}$ j\d{e}$_{\text{yo}}$ Yoruba$_{\text{yo}}$ word$_{\text{en}}$.
		
		\textit{I said you should relax, infact I accept the surprise that spreadsheet is a Yoruba word.}
		};
	\end{tikzpicture}
	
	\vspace{1mm}
	\textbf{Topic}: social media; \textbf{keyword}: cope; \textbf{Pronoun}: indefinite; \textbf{Tense}: present; \textbf{Use negative particle}: Yes\\[1ex]
	\begin{tikzpicture}
		\node[text width=0.97\columnwidth, rounded corners, fill=blue!10, draw=none, anchor=north] {
			\textbf{Example 2}: K\`o$_{\text{yo}}$ s\'i$_{\text{yo}}$ \`e\`ey\`an$_{\text{yo}}$ t\'o$_{\text{yo}}$ y\`an$_{\text{yo}}$ \`on\`a$_{\text{yo}}$ n\'i$_{\text{yo}}$ w\'ah\'al\`a$_{\text{yo}}$, view$_{\text{en}}$ y\`i\'i$_{\text{yo}}$ ni$_{\text{yo}}$ aw\d{o}n$_{\text{yo}}$ \d{\`e}d\'a$_{\text{yo}}$ t\'i$_{\text{yo}}$ w\d{\`o}n$_{\text{yo}}$ \d{s}e$_{\text{yo}}$ l\`ati$_{\text{yo}}$ cope$_{\text{yo}}$. 
			
			\textit{There is no person that chooses problems as a path, this view is what the creatures XXX did to cope}			
		};
	\end{tikzpicture}
	
		Examples 1 and 2 both follow the prompt guidelines with respect to the matrix language and tense. Example 1, however, uses a personal pronoun instead of an indefinite pronoun with Example 2 using the correct pronoun. XXX in Example 2 indicates a phrase that cannot be translated.
		
		We observe that the prompting approach can also support the generation of Yoruba--English sentences that are diverse. 
		
		We provide observations on the coherence and naturalness of synthetic sentences in Section~\ref{sec:evaluation:langspes}.

%% file: evaluation.tex
\section{Evaluation of Generated Data}
\label{sec:evaluation}

In this section,  we evaluate our work in three parts: (i) we evaluate the diversity of the generated sentences, (ii) we comment on GPT 3.5's adherence to the prompts provided, and (iii) we evaluate the quality of the sentences generated through a combination of statistical analysis and human evaluation of the sentences. We use the four prompt guidelines as discussed in Section~\ref{sec:method}. For this paper we Romanised the Yoruba--English sentences for easier evaluation, however, we will include this in future work. 

\subsection{Data Diversity}
\label{sec:evaluation:div}
\subsubsection{Content Diversity}
\begin{figure*}
	\centering
	\begin{subfigure}[b]{0.45\textwidth}
		\centering
 		\includegraphics[width=\textwidth]{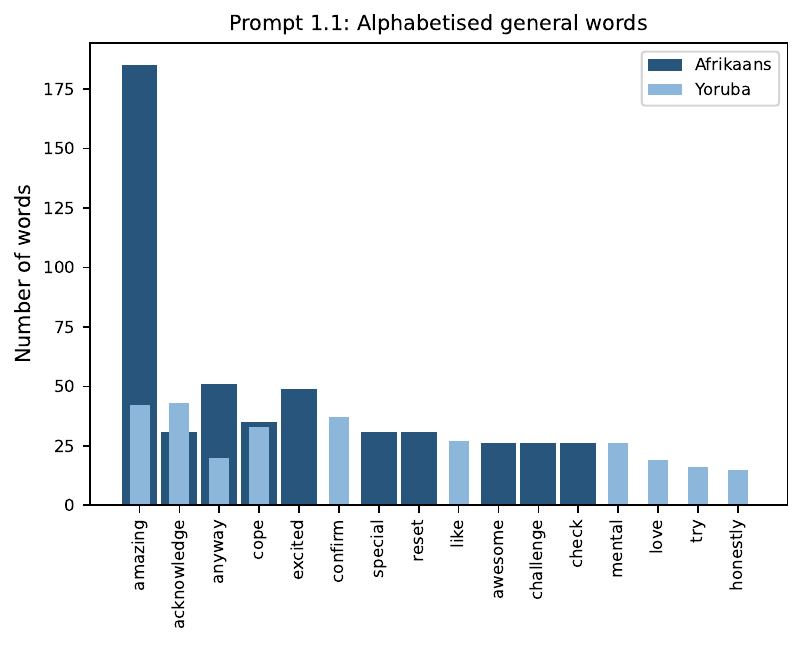}
		\caption{Distribution of general words (alphabetised).}
		\label{fig:gen_exp2}
	\end{subfigure}
	\hfill
	\begin{subfigure}[b]{0.45\textwidth}
		\centering
		\includegraphics[width=\textwidth]{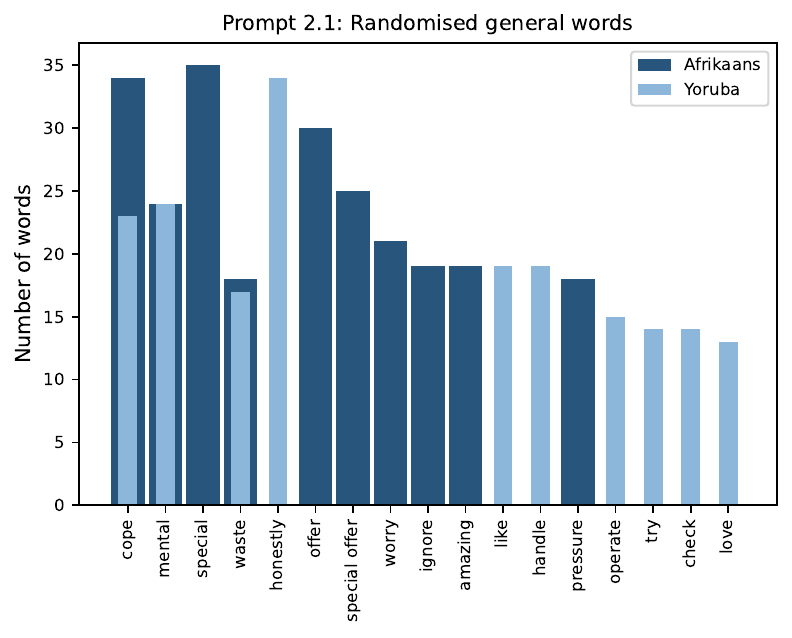}
		\caption{Distribution of general words (randomised).}
		\label{fig:gen_exp9}
	\end{subfigure}
	
	\caption{Distribution of top 10 general CS words across all topics.}
	\label{fig:gen_words}
\end{figure*}

 In Figure~\ref{fig:gen_exp2} (from Prompt 1.1) we see a large amount of general words being used compared with the number of sentences. We also note that the top three keywords (\textit{amazing, acknowledge, anyway}) is the same as the top three keywords in the alphabetised list. In Prompt 2.1 we provide a randomised general word list to GPT 3.5 and in Figure~\ref{fig:gen_exp9} we observe a more even distribution of general words as a result. This indicates GPT 3.5's sensitivity to prompts and the context provided. 

\subsubsection{Linguistic Diversity}

Since Prompts 2.1 and 2.2 asked ``start the sentence with...'', all sentences were evaluated accordingly. We used a list of common Afrikaans and Yoruba pronouns to evaluate this prompt.
 
\begin{figure}
	\centering
	\hspace*{-0.4cm}
		\includegraphics[width=3in]{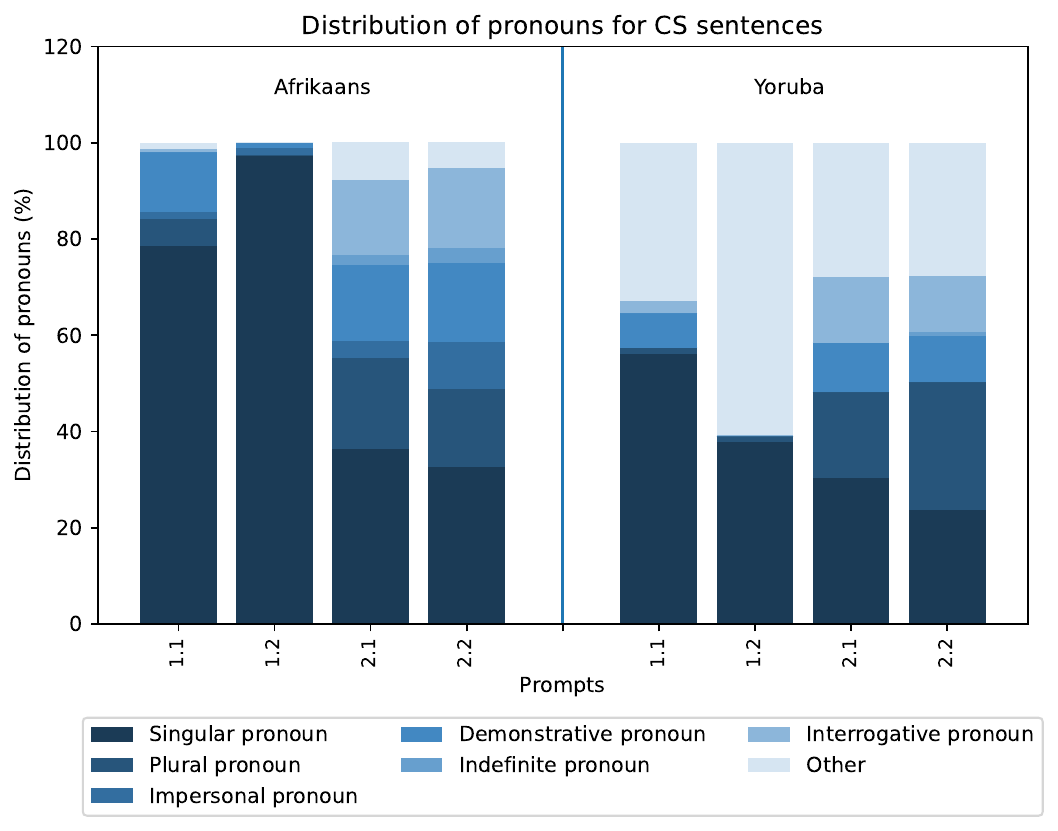}
		\caption{Distribution of pronouns.}
		\label{fig:afr_pron}
		
	\caption{Distribution of the use of pronouns at the beginning of a generated sentence.}
	\label{fig:pron_dist}
\end{figure}

From Figure~\ref{fig:pron_dist} we observe an increase in diversity of the types of sentences with regards to the distribution of pronouns (Prompts 2.1/2.2). For Afrikaans--English, more than 90\% of the sentences start with one of the specified pronouns. We also see an increase in the diversity of Yoruba--English sentences, however, there are still \mbox{$\sim$35\%} of sentences starting with words other than the requested pronouns. It is not well understood why GPT 3.5 ignored these prompts. In the absence of linguistic guidelines in the prompt, we note that by adding few-shot examples, we lack diversity (Prompts 1.2 and 2.2).
 
Similarly to pronouns, we use Afrikaans and Yoruba keywords that indicate past and future tense, negation (negative sentiment) and conjunctions to evaluate the effect of adding these guidelines to the prompts. In Table~\ref{dist_oth_prompt} we highlight the impact of these factors on distribution in sentences using Prompts 1.1 and 2.1 (prompts without example sentences).
\begin{table}[ht]
	
	\begin{center}
	\begin{tabularx}{\columnwidth}{lp{0.8cm}p{0.8cm}p{0.8cm}p{0.8cm}}
			\toprule
			&\multicolumn{2}{c}{\textbf{Afrikaans}}&\multicolumn{2}{c}{\textbf{Yoruba}}\\
			Prompt&\textbf{1.1}&\textbf{2.1}&\textbf{1.1}&\textbf{2.1}\\
			\midrule
			Past Tense&42\%&34\%&17\%&23\%\\
			Future Tense&55\%&39\%&10\%&12\%\\			
			Negation&26\%&39\%&15\%&27\%\\
			Conjunction*&14&4&1&2\\
			\bottomrule
			\multicolumn{5}{l}{\footnotesize *The ratio of Afrikaans/Yoruba to English.}\\	
	\end{tabularx}
		
		\caption{Distribution of tenses and negation and ratio of conjunctions.}
		\label{dist_oth_prompt}
	\end{center}
	
\end{table}

We see in Table~\ref{dist_oth_prompt} that for Afrikaans--English, both the distribution of tenses (equal distribution between past and future) and the presence of negation improved. However, it is only negation that improved for Yoruba--English. We further elaborate on this observation in Section~\ref{sec:evaluation:stats}. The ratio of Afrikaans:English conjunctions decreased showing the guideline is not efficient. For Yoruba:English conjunctions we observe a slight improvement.

The above statistical evaluation of diversity shows that adding various linguistic guidelines to the prompts improves diversity. However, this does not consider whether a prompt is adhered to. In the next section, we evaluate GPT 3.5's ability to execute prompts.

\subsection{Prompt Adherence}
In Section~\ref{sec:dataset:prompt} we already observed that GPT 3.5 does not always adhere to using the specified matrix language and since we do not consider word-level language identification in this paper, we exclude this when determining adherence. 

We apply a simple approach to calculate prompt adherence. We express the number of prompts adhered to as a percentage of the total prompts given. In Prompt 1.1, the only prompt given is the topic keyword hence a total of one prompt (the same for Prompt 1.2). In Prompt 2.1, there are five prompts given: topic keyword, pronoun, tense, negative particle and conjunction. The average prompt adherence across the sentences is then used to represent overall prompt adherence.

\subsubsection{Statistical Evaluation of Prompt Adherence}
\label{sec:evaluation:stats}

In this section we present the prompt adherence for the four prompt guidelines. Keywords for pronouns, tenses, negative particles and conjunctions as per Section~\ref{sec:evaluation:stats}. Table~\ref{tbl:overall_adhere} shows the overall prompt adherence.
 
\begin{table}[ht]
	
	\begin{center}
		\begin{tabular}{lp{0.8cm}p{0.8cm}p{0.8cm}p{0.8cm}}
			
			\toprule
			\textbf{Prompt}&\textbf{1.1}&\textbf{1.2}&\textbf{2.1}&\textbf{2.2}\\
			
			\midrule
			Afrikaans&83\%&90\%&74\%&78\%\\
			Yoruba&83\%&92\%&53\%&58\%\\	
			\bottomrule
		\end{tabular}
		\caption{Overall prompt adherence.}
		\label{tbl:overall_adhere}
	\end{center}
\end{table}
From Table~\ref{tbl:overall_adhere} we see that the adherence to prompts for Yoruba--English is much lower than for Afrikaans--English in the linguistically guided prompts (Prompts 2.1 and 2.2). 

In Afrikaans there are a few specific keywords such as \textbf{`nie', `nooit', `nee'} \textit{(English: not, never, no)} that indicate negation. Similarly for tenses, words like \textbf{`was', `gister', `wil', `more'} \textit{(English: was, yesterday, will, tomorrow)} can be used for past and future tense. However, the Yoruba language is more complex and keywords like the above-mentioned are not adequate to identify negation and tenses, hence the lower prompt adherence.

In the next section (Section~\ref{sec:evaluation:manual}) we use manual annotation of sentences for tenses and negation to re-evaluate prompt adherence. 

\subsubsection{Manual Evaluation of Prompt Adherence}
\label{sec:evaluation:manual}
For manual evaluation of generated sentences, we sample 100 sentences each from the four prompt methods.

We manually annotate the sentences of Prompts 2.1 and 2.2 with tense (past or future) and negation (whether the sentence expresses some negative sentiment). In future work, external annotators will also be used.

In Table~\ref{upd_prompt} we show the impact on the calculated prompt adherence (using Prompt 2.1) for the statistical (1) and manual (2) evaluation of the 100 sentences. The prompt adherence for Yoruba--English increased to 66\% from 59\% with a significant increase in the adherences to tenses.Afrikaans--English prompt adherence remains constant. The adherence to negation reduced slightly for both languages. This confirms the earlier comment that it is statistically more difficult to calculate prompt adherence for Yoruba--English without a human in the loop.

\begin{table}[ht]
	
	\begin{center}
		\begin{tabularx}{\columnwidth}{lp{1cm}p{1cm}p{1cm}p{1cm}}
			\toprule
			&\multicolumn{2}{c}{\textbf{Afrikaans}}&\multicolumn{2}{c}{\textbf{Yoruba}}\\
			Prompt&\textbf{(1)}&\textbf{(2)}&\textbf{(1)}&\textbf{(2)}\\
			\midrule
			Tense&79\%&84\%&41\%&72\%\\		
			Negation&47\%&41\%&40\%&36\%\\
			\textbf{Total}&72\%&72\%&59\%&66\%\\
			\bottomrule
			
		\end{tabularx}
		
		\caption{Comparing prompt adherence for both a statistical and manual annotation perspective.}
		\label{upd_prompt}
	\end{center}
\end{table}

We conclude that there is potential in using GPT 3.5 as a supporting tool to generate diverse sentences with linguistically guided prompts. In the following sections we provide an overview of the quality of generated sentences to further determine the role that GPT 3.5 can play in addressing code-switched data availability. 

\subsection{Code-Switch Acceptability}
\label{sec:evaluation:manual}

The final part of our analysis looks at the quality of generated sentences. As mentioned in Section~\ref{sec:evaluation:manual}, we sampled 100 sentences from each of the four prompt methods. For this part of the analysis, we rated the acceptability of a code-switch sentence according to: i) Yes, ii) Yes, with minimal changes or iii) No. We adopt the constraint-free approach of \citet{macswan2000architecture}.
 
\begin{figure}[!ht]
	\begin{center}
		\hspace*{-0.4cm}
		\includegraphics[width=3in]{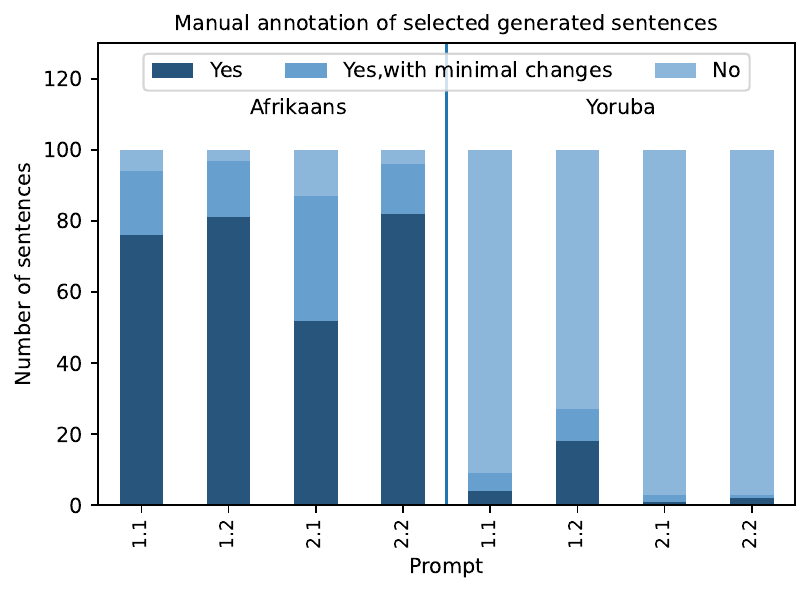} 
		
		\caption{Evaluation of manual annotation of sentences.}
		\label{manual}
	\end{center}
\end{figure}

The results of the manual annotation are shown in Figure~\ref{manual}. We observe that the acceptability of Afrikaans--English sentences far outweighs that of Yoruba--English. We also see that adding few-shot examples increases acceptability (Prompts 1.2 and 2.2). Although we observe an increase in diversity through linguistic guidelines, the quality of sentences are sub-optimal. Subsequent work will focus on how correctable sentences can be used for improved prompting and/or fine tuning of language models. However, with further analysis and improvement, there is potential to use GPT 3.5 to support synthetic data generation.

\subsection{Language Specific Observations}
\label{sec:evaluation:langspes}

\subsubsection{Afrikaans--English}

In order to quantify the acceptability observed from internal evaluation, we randomly select 5 Afrikaans-English sentences from the dataset used for manual evaluation (Section~\ref{sec:evaluation:manual}). Table~\ref{tbl:afr_analysis} gives the sentences with translations and comments.

\begin{table*}[!ht] 
	\centering
	\small
	\renewcommand{\arraystretch}{0.1}  
	\begin{tabularx}{\textwidth}{cp{0.45\textwidth}cp{0.38\textwidth}}
		\toprule
		& Sentence & Accept & Comments \\
		\midrule
		\ii{1} & Ek is so excited om my nuwe partner te ontmoet. \textit{(I am so excited to meet my new partner.)}& \checkmark & -\\\\
		\addlinespace
		\ii{2} & Ons moet takeaways hê for dinner, maar ek wil nie weer McDonald's eet nie. \textit{(We must have takeaways for dinner, but I don't want to eat McDonald's again.)}& \checkmark & The use of English `for' instead of Afrikaans `vir' is less typical but can be accepted\\\\
		\addlinespace
		\ii{3} & Ons het 'n nuwe app gedownload om die fotos te organise. \textit{(We downloaded a new app to organise the photos.)}& \checkmark & `gedownload' is an example of intra-word code-switching\\\\
		\addlinespace
		\ii{4} & Ek moet 'n nuwe uitdaging in my loopbaan aanpak. \textit{(I have to tackle a new challenge in my career.)}& \xmark & No code-switching, only Afrikaans\\\\
		\addlinespace
		\ii{5} & Daai kursus was 'n disaster, ons het reset van die begin af. \textit{(That course was a disaster, we reset from the beginning.)}& \xmark & Unclear about the intended meaning with the use of `reset', however, can be corrected in context\\\\
		\addlinespace
		\bottomrule
		\end{tabularx}
		\caption{Generated Afrikaans--English sentences, translations and comments on acceptability.}
		\label{tbl:afr_analysis}
	\end{table*}

In our general overview we find that the typical mistakes made are as a result of following English grammar structure. However, for many sentences this does not affect the meaning and can be corrected. 

The results from the various experiments therefore indicate that using GPT 3.5 (and it's followers) can be considered as a method to generate large-scale data in Afrikaans-English code-switching.

\subsubsection{Yoruba--English}

Similarly to quantifying the Afrikaans-English sentences, we give 5 randomly selected Yoruba-English sentences in Table~\ref{tbl:yoruba_analysis} from the dataset used for manual evaluation (Section~\ref{sec:evaluation:manual}).

\begin{table*}[!ht] 
	\centering
	\small
	\renewcommand{\arraystretch}{0.1}  
	\begin{tabularx}{\textwidth}{cp{0.45\textwidth}cp{0.38\textwidth}}
		\toprule
		& Sentence & Accept & Comments \\
		\midrule
		\ii{1} & o ma install software yii ni computer mi. \textit{(You will install this software in my computer.)}& \checkmark & The model is not clear about the right orthography for the Yoruba words in the sentence and used the word ``ni" instead of `sorii' which translates to `on' in Yoruba\\\\
		\addlinespace
		\ii{2} & 60 million naira yen fe po die fun mi. I need to buy orange juice for the party. \textit{(That 60 million naira seems to be a bit too much for me. I need to buy orange juice for the party.)}& \checkmark & This is an inter-sentential code-switched sentence. However, this can be accepted by just dropping the second sentence\\\\
		\addlinespace
		
		\ii{3} & Mo n gbadun ojo meta ti n si se fun mi ni lockdown ni ojo kan, but honestly, e wa wo mi, I don tire. The pressure don too much, and I just dey try survive. \textit{(I am enjoying the three days XXX during lockdown in one day, but honestly, come and see me, I am tired.)} & \xmark\ & These sentences make no sense. Contains the Nigerian version of Pidgin-English mixed with Yoruba and English. The `XXX' indicates phrases that cannot be translated\\\\
		\addlinespace
		
		\ii{4} & eniyan miran naa maa click si awon idile mi lati ba wa. \textit{(That other person will click to my family to come with.)}& \xmark & This sentence makes no sense  \\\\
		\addlinespace
		
		\ii{5} & 	o ma jabo ile-ise yi niwaju wireless connectivity yi.  \textit{(You will XXX this company in front of this wireless connectivity.)} & \xmark & The English translation for the word `jabo' cannot be inferred  without knowing the diacritics. The sentence makes no sense \\\\
		\addlinespace
	
		\addlinespace
		\bottomrule
	\end{tabularx}
	\caption{Generated Yoruba--English sentences, translations and comments on acceptability.}
	\label{tbl:yoruba_analysis}
\end{table*}

It is hypothesised that the exposure of GPT 3.5 to the Yoruba language is to a much lesser extent than Afrikaans yielding a substantial amount of unacceptable sentences. Furthermore, as was postulated by \citet{Yong2023}, languages using the English alphabet and Latin script perform better on LLMs. Further analysis is required to improve prompting and quality of sentences.

%% file: conclusions.tex
\section{Conclusions and Future Work}
\label{sec:conclusions}

In this paper we extended on the of \citet{Yong2023} where they used prompting of LLMs (including GPT 3.5) to generate code-switch sentences. Our approach evaluates three dimensions: (i) diversity, through a wider range of topics, keywords, linguistic guidelines and few-shot examples; (ii) prompt adherence, to understand the ability of GPT 3.5 to follow these prompts; and (iii) quality, to determine the use of GPT 3.5 as a supporting tool to address code-switched data scarcity.
We evaluated two typologically diverse language pairs: Afrikaans--English and Yoruba--English. 

Our main findings are: (i) using topics, keywords and general context words increases coverage; (ii) linguistic-based guidelines increases diversity in the types of sentences, (iii) few-shot prompting increases the quality of sentences but is limited in diversity of the types of sentences;(iv) quality of sentences are much lower for languages that use non-Latin script (such as Yoruba); and (v) evaluating quality of data requires a human-in-the-loop.
We provide a framework for linguistically-guided prompting and we conclude that \textit{OpenAI's GPT} exhibits the ability to support synthetic code-switched data generation and can be invaluable to address the issue of data availability.

In future work we will address the following: i) include external annotation to cross-validate the quality of generated sentences; (ii) improve on the prompting guidelines to increase quality; (iii) use correctable sentences to improve the performance of the latest generation of \textit{OpenAI's GPT} to support large-scale generation; and (iv) expand to more African languages in an effort to develop a language agnostic approach to synthetically generate data.

\section{Ethical Considerations}
\paragraph{Data Generation} Research in code-switching is not only focused on the grammatical aspects of this phenomenon but also the socio-pragmatic characteristics in discourse \citep{Nel2012}. Large language models such as \textit{OpenAI's GPT} are influenced by social views and inherit encoded biases \citep{Bender2021}. Our work propose the use of GPT to support efforts in synthetically generated code-switched data to increase the prevalence of under-resourced languages. We therefore carefully considered the method in which GPT was prompted to eliminate the introduction of bias. We use general topics and keywords with the goal to generate a diverse range of acceptable sentences.
\paragraph{Human Evaluation} The generated sentences were internally evaluated by native speakers of Afrikaans and Yoruba. We ensure the data is respectful to culture and social norms. We will continue to include humans-in-the-loop to ensure faithful data generation.

\section{Acknowledgements}
We thank JP Morgan and ABSA for their financial support, and OpenAI for providing API credits.